\documentclass[twocolumn,10pt, cleanfoot]{asme2ej}

\usepackage{graphicx} %% for loading jpg figures
\usepackage{xcolor}

\usepackage[T1]{fontenc}
\usepackage{tikz}
\usetikzlibrary{scopes}
\usetikzlibrary{%
    decorations.pathreplacing,%
    decorations.pathmorphing%
}

\usepackage{mathptmx} % assumes new font selection scheme installed
\usepackage{amsmath} % assumes amsmath package installed
\usepackage{amssymb}  % assumes amsmath package installed
\usepackage{float}
\usepackage{algorithm}
\usepackage[noend]{algpseudocode}
\usepackage{subcaption}
\usepackage{booktabs}

\title{Adaptive Identification of Legged Robotic Kinematic Structure}

\author{Bolun Dai
    \affiliation{
	Department of Mechanical Engineering\\
	Carnegie Mellon University\\
	Pittsburgh, PA 15217\\
    bdai@andrew.cmu.edu
    }	
}

\begin{document}
\maketitle
%%%%%%%%%%%%%%%%%%%%%%%%%%%%%%%%%%%%%%%%%%%%%%%%%%%%%%%%%%%%%%%%%%%%%%
\begin{abstract}
{\it
Model-based control usually relies on an accurate model, which is often obtained from CAD and actuator models. The more accurate the model the better the control performance. However, in bipedal robots that demonstrate high agility actions, such as running and hopping, the robot hardware will suffer from impacts with the environment and deform in vulnerable parts, which invalidates the predefined model. Thus, it is desired to have an adaptable kinematic structure that takes deformation into consideration. To account for this we propose an approach that models all of the robotic joints as 6-DOF joints and develop an algorithm that can identify the kinematic structure from motion capture data. We evaluate the algorithm's performance both in simulation -- a three link pendulum, and on a bipedal robot -- ATRIAS. In the simulated case the algorithm produces a result that has a $3.6\%$ error compared to the ground truth, and on the real life bipedal robot the algorithm's result confirms our prior assumption where the joint deform on out-of-plane degrees of freedom. In addition our algorithm is able to predict torque and forces using the reconstructed joint mode.
}
\end{abstract}
%%%%%%%%%%%%%%%%%%%%%%%%%%%%%%%%%%%%%%%%%%%%%%%%%%%%%%%%%%%%%%%%%%%%%%
\section{INTRODUCTION}

Model-based control is ubiquitous in legged robotic systems \cite{DBLP:conf/robio/DesaiG12}, \cite{DBLP:conf/iros/RutschmannSBB12}, \cite{Martin-2015-102704}, \cite{7354279}. Given that legged robotic systems are more complex compared to robotic arms and wheeled robots in controller structures, implementing model-based control helps to achieve close-to-optimal performance and also improves robustness while providing a physical insight of the system. Because of the many benefits of model-based control various teams in the DARPA Robotics Challenge implemented model-based control on their humanoid robots \cite{DBLP:journals/jfr/FengWXA15}, \cite{7029950}, \cite{7363480}, \cite{doi:10.1002/rob.21685}, \cite{doi:10.1002/rob.21571}. However, model-based control relies on a specific kinematic model and its correlated dynamic parameters, no matter explicit or implicitly given. Without an accurate enough mathematical model of the system, model-based control will be less effective. Work has been done in identifying the dynamic parameters of a robotic system \cite{DBLP:journals/nn/TingDS11}, \cite{DBLP:journals/jr/DingWYY15}, \cite{DBLP:conf/icra/GautierJV08}. However, little has been done in identifying the kinematic parameters of legged robotics systems. Current works mostly focuses on identifying the manipulator stiffness of industrial robotic arms \cite{ABELE2007387}, \cite{DUMAS2011881}, \cite{doi:10.1177/1687814018761297} which are much more rigid compared to legged robots. In addition, work has been done in identifying flexible joint stiffnesses on the Canadarm \cite{doi:10.2514/1.G000197}. This paper aims to extend the work on to legged robots.

Kinematic models of robots are often presumed from CAD models and actuator models. The underlying assumption for most model-based control methods is that they do not change over time. However, in reality the robot themselves often shows bending and twisting during movement. One example is the ATRIAS bipedal robot \cite{doi:10.1177/0278364916648388}, when experimenting with its ability in locomotion it is observed that the knee joint experiences rotation about axis other than the designed rotation axis, this rotation is especially large when demonstrating running gaits or hopping. The legs of ATRIAS consists of four links which constructs a plane, rotation about any axis at the joints that is not perpendicular to the leg plane will deform the legs. Such deformation will produce a force that is out of the leg plane which results in a torque that causes the torso of ATRIAS to tilt, which eventually leads to the robot falling down due to large pitch or roll movements. Thus, the mismatch between the suggested kinematic model and the actual kinematic model may lead to serious issues.

It would be preferable for identifying the kinematic structure of the robot adaptively by observing the movement. For model-based control to be more stable and effective we need to develop novel methods in obtaining a more accurate model of the robotic system. One approach for improving the modelling accuracy is to make more generalized assumptions. Work has been done in identifying actuator models using a data driven approach to capture the nonlinear effects \cite{Hwangboeaau5872}. This paper presents an approach to get a more accurate kinematic model by assuming that all robot joints are essentially 6 degrees-of-freedom (DOF) joints instead of the widely used prismatic joints and revolute joints, and proposed a model for the interaction inside the joint. Using this model we can take into account the deformation of robotic joints during high impact movements such as running and hopping.

This remainder of this paper is organized into three parts. In section II, We first provide an overview of the joint model we proposed, the kinematic equations of the links when using the proposed joint model and the corresponding method for identifying the parameters of the joint model. Then we show results for the joint parameter identification in simulation. In section III we present the results for the joint parameter identification on ATRIAS biped. And in section IV we will discuss the effectiveness of the model and future directions.
\section{APPROACHES}

This section is structured as the follow, first the model of a 6 DOF joint will be given, then the kinematic structure of a robotic link with 6 DOF joints will be shown, then a joint parameter reconstruction algorithm will be presented, finally the simulation result of the reconstruction algorithm will be demonstrated.
\subsection{Joint Model}
We proposed a robotic joint model that takes the deformation of robotic joints during movement into consideration. A joint connects two links, the two links can move relatively depending on the the type of the joint. The movement of the two links is modeled as the movement of two frames: the base frame and the follower frame. Unlike a traditional prismatic joint, where the follower frame only moves along an axis defined in the base frame, or a revolute joint, where the follower frame can only rotate about an axis defined in the base frame, we make no constrains on the relative movement between the base and follower frame in our joint model. This means that the links that are connected by a 6DOF joint can move relatively on six DOF, three for translation and three for rotation. The movement of each DOF is governed by a spring-damper system and we can mathematically model the movement of the follower frame relative to the base frame as
\begin{equation}
	\begin{cases}
		F_x = k_{px}\Delta{x} + k_{dx}\dot{x} & \\
		F_y = k_{py}\Delta{y} + k_{dy}\dot{y} & \\
		F_z = k_{pz}\Delta{z} + k_{dz}\dot{z} & 
	\end{cases}
	\label{eq:f}
\end{equation}
\begin{equation}
	\begin{cases}
		\tau_x = k_{p\theta x}\Delta{\theta_x} + k_{d\theta x}\dot{\theta_x}\\
		\tau_y = k_{p\theta y}\Delta{\theta_y} + k_{d\theta y}\dot{\theta_y}\\
		\tau_z = k_{p\theta z}\Delta{\theta_z} + k_{d\theta z}\dot{\theta_z}
	\end{cases}
	\label{eq:t}
\end{equation}
where equation \ref{eq:f} governs the translational movement of the follower frame relative to the base frame. The $k_{px}$, $k_{py}$ and $k_{pz}$ represents the translational spring stiffness along the $x$, $y$ and $z$ axis, the $k_{dx}$, $k_{dy}$ and $k_{dz}$ represents the translational damping coefficient. Similarly equation \ref{eq:t} governs the rotational movement of the follower frame relative to the base frame, with $k_{p\theta x}$, $k_{p\theta y}$ and $k_{p\theta z}$ representing the rotational spring stiffness and $k_{d\theta x}$, $k_{d\theta y}$ and $k_{d\theta z}$ representing the rotational damping coefficient. For a 6DOF joint given that the original design of a robot joint is not to let it freely move along  6 degrees of freedom, therefore we can see that using such a method we can see that the spring stiffness and damping coefficient among different degrees of freedom varies greatly. For those degrees of freedom that the joint is originally designed to have motion we can observe a low spring stiffness and damping coefficient, we say that this is the actuator DOF. And for some degrees of freedom the spring stiffness and damping coefficient is higher than the actuator DOF, but smaller than those that observe little movement, we say these are the unexpected movement. And for those DOF that have really high spring stiffness and damping coefficient we say that they are rigid movement. Note that since both equation \ref{eq:f} and \ref{eq:t} describes the movement of the follower frame relative to the base frame, the $x$, $y$ and $z$ axis makes up the body frame which aligns with the base frame. Using the aforementioned joint model we can describe the motions of the robotic joints on DOF's that are other than its designed DOF, which is a common phenomenon in legged robotics.

\subsection{Kinematic Structure}

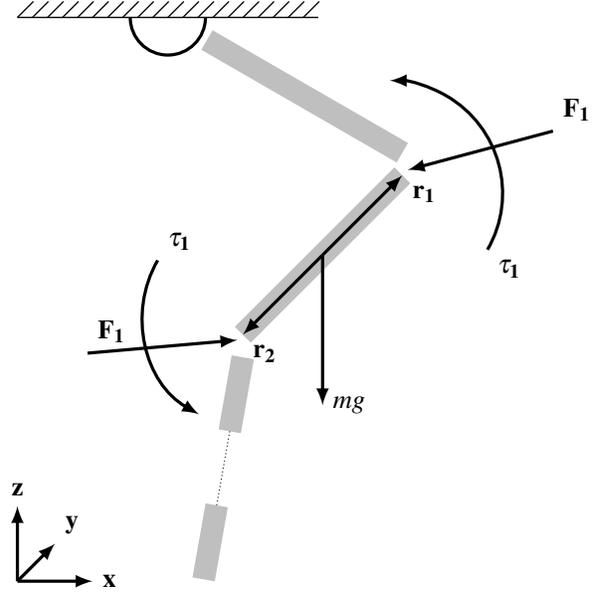
\begin{figure}
    \centering
    \begin{tikzpicture}
    [
    force/.style={>=latex,draw=blue,fill=blue},
    interface/.style={
        % The border decoration is a path replacing decorator. 
        % For the interface style we want to draw the original path.
        % The postaction option is therefore used to ensure that the
        % border decoration is drawn *after* the original path.
        postaction={draw,decorate,decoration={border,angle=45,
                    amplitude=0.3cm,segment length=2mm}}},
    ]
    % \draw[step=0.25cm,color=gray] (-1,-1) grid (1,1);
    % Draws the interface along with the dashed lines
    \path (-2, -7.5) coordinate (coordinate_origin);
    \draw[very thick, ->, >=latex] (coordinate_origin) -- ++(0, 1)node[above] {$\mathbf{z}$}; 
    \draw[very thick, ->, >=latex] (coordinate_origin) -- ++(1, 0)node[right] {$\mathbf{x}$}; 
    \draw[very thick, ->, >=latex] (coordinate_origin) -- ++(45:0.707)node[above right] {$\mathbf{y}$};

    \draw[black,line width=.5pt,interface](-2,0)--(2,0);
    % Draws the arc at the origin
    \draw[very thick] (0.5,0) arc (0:-180:0.5cm);
    % Draws the first link
    \def\ang1{-30}
    \path (\ang1:0.6cm) coordinate (Link1Head);
    \path (Link1Head) ++(\ang1:3cm) coordinate (Link1Tail);
    \path (Link1Head) ++(\ang1-90:0.15cm) ++(0, 1cm) coordinate (up1);
    \path (Link1Tail) ++(\ang1+90:0.15cm) ++(0, -1cm) coordinate (down1);
    
    \begin{scope}
        \clip (Link1Head) -- ++(\ang1+90:0.15cm) 
                          -- ++(\ang1:3cm) 
                          -- (Link1Tail) 
                          -- ++(\ang1-90:0.15cm)
                          -- ++(\ang1+180:3cm)
                          -- cycle;
        \fill[lightgray] (up1) rectangle (down1);
    \end{scope}
    
    % Draws the second link
    \def\ang2{-135}
    \path (Link1Tail) ++(0, -0.3) coordinate (Link2Head);
    \path (Link2Head) ++(\ang2:3cm) coordinate (Link2Tail);
     \path (Link2Head) ++(\ang2:1.5cm) coordinate (Link2CoM);
    \path (Link2Head) ++(\ang2+90:0.15cm) ++(0, 1cm) coordinate (up2);
    \path (Link2Tail) ++(\ang2+90:0.15cm) ++(-1, 0cm) coordinate (down2);
    \path (Link2Head) ++(\ang2+180:0.1cm) coordinate (force1head);
    \path (force1head) ++(15:2cm) coordinate (force1tail);
    \path (Link2Tail) ++(\ang2:0.1cm) coordinate (force2head);
    \path (force2head) ++(-175:2cm) coordinate (force2tail);
    
    \begin{scope}
        \clip (Link2Head) -- ++(\ang2+90:0.15cm) 
                      -- ++(\ang2:3cm) 
                      -- (Link2Tail) 
                      -- ++(\ang2-90:0.15cm)
                      -- ++(\ang2+180:3cm)
                      -- cycle;
        \fill[lightgray] (up2) rectangle (down2);
    \end{scope}
    
    \draw[very thick, ->, >=latex] (force1tail)node[above right] {$\mathbf{F_1}$} -- (force1head);
    \draw[very thick, ->, >=latex] (force1head) ++(\ang2+90:1.5cm)node[below right] {$\mathbf{\tau_1}$} arc (-30:90:1.5cm);
    \draw[very thick, ->, >=latex] (force2tail)node[above right] {$\mathbf{F_1}$} -- (force2head);
    \draw[very thick, ->, >=latex] (force2head) ++(\ang2-90:1.5cm)node[above right] {$\mathbf{\tau_1}$} arc (150:240:1.5cm);
    \draw[very thick, ->, >=latex] (Link2CoM) -- ++(0, -2)node[right] {$mg$};
    \draw[very thick, ->, >=latex] (Link2CoM) -- (Link2Head)node[below right] {$\mathbf{r_1}$};
    \draw[very thick, ->, >=latex] (Link2CoM) -- (Link2Tail)node[below right] {$\mathbf{r_2}$};
    
    % Draws the third link
    \def\ang3{-100}
    \path (Link2Tail) ++(0, -0.3) coordinate (Link3Head);
    \path (Link3Head) ++(\ang3:1cm) coordinate (Link3Tail);
    \path (Link3Head) ++(\ang3+90:0.15cm) ++(0, 1cm) coordinate (up3);
    \path (Link3Tail) ++(\ang3+90:0.15cm) ++(-1, 0cm) coordinate (down3);
    
    \begin{scope}
        \clip (Link3Head) -- ++(\ang3+90:0.15cm) 
                          -- ++(\ang3:1cm) 
                          -- (Link3Tail) 
                          -- ++(\ang3-90:0.15cm)
                          -- ++(\ang3+180:1cm)
                          -- cycle;
        \fill[lightgray] (up3) rectangle (down3);
    \end{scope}
    
    % Draws the n-th link
    \def\ang4{-100}
    \path (Link3Tail) ++(\ang4:1cm) coordinate (LinknHead);
    \path (LinknHead) ++(\ang4:1cm) coordinate (LinknTail);
    \path (LinknHead) ++(\ang4+90:0.15cm) ++(0, 1cm) coordinate (upn);
    \path (LinknTail) ++(\ang4+90:0.15cm) ++(-1, 0cm) coordinate (downn);
    
    \begin{scope}
        \clip (LinknHead) -- ++(\ang4+90:0.15cm) 
                          -- ++(\ang4:1cm) 
                          -- (LinknTail) 
                          -- ++(\ang4-90:0.15cm)
                          -- ++(\ang4+180:1cm)
                          -- cycle;
        \fill[lightgray] (upn) rectangle (downn);
    \end{scope}
    
    \draw[densely dotted] (Link3Tail) -- (LinknHead);

\end{tikzpicture}
    \caption{This graph shows how the parameters in equation \ref{eq:sum1} are defined, with $r_1$ being the distance between the center-of-mass (CoM) of the link and the follower frame of the head joint where $F_1$ is applied at, similarly $r_2$ is the distance between the CoM and the base frame of the tail joint where $F_2$ is applied at. $\tau_1$ and $\tau_2$ are the torques that are applied on the head and tail joint, $\tau_1$ is applied at the same end as $F_1$, the same goes for $\tau_2$ and $F_2$. All of these parameters are given using the world coordinate.}
\label{fig:kine_struct}
\end{figure}

Given the 6 DOF joint model, we need to define the movement of robotic links in terms of the joint parameters and measurable or already known link parameters. First, we need to clarify the conventions that will be used here. A typical robotic linkage systems will be serially positioned, one can pre-define a head and a tail for the whole linkage system. Following such a guidance we can define the relation between a specific link and the links that it is connected with: we call the link that is before the link in interest in such definition as the previous link, the link in interest as the current link and the following link as the next link. Also we have the joint that connects the current link with the previous link as the head joint and the link that connects the current link and the next link as the tail joint. We assume that the link itself has forces and torques applied on both ends of the link. Using the Newton-Euler equation we can get
\begin{equation}
	r_1\times F_1 + r_2\times F_2 + \tau_1 + \tau_2 = \sum{\tau}
	\label{eq:sum1}
\end{equation}
\begin{figure}
    \centering
    \begin{tikzpicture}
    [
    media/.style={font={\footnotesize\sffamily}},
    wave/.style={
        decorate,decoration={snake,post length=1.4mm,amplitude=2mm,
        segment length=2mm},thick},
    force/.style={>=latex,draw=blue,fill=blue},
    interface/.style={
        % The border decoration is a path replacing decorator. 
        % For the interface style we want to draw the original path.
        % The postaction option is therefore used to ensure that the
        % border decoration is drawn *after* the original path.
        postaction={draw,decorate,decoration={border,angle=45,
                    amplitude=0.3cm,segment length=2mm}}},
    ]
    % \draw[step=0.25cm,color=gray] (-1,-1) grid (1,1);
    % Draws the interface along with the dashed lines
    \path (-0, -6) coordinate (coordinate_origin);
    \draw[very thick, ->, >=latex] (coordinate_origin) -- ++(0, 1)node[above] {$\mathbf{z}$}; 
    \draw[very thick, ->, >=latex] (coordinate_origin) -- ++(1, 0)node[right] {$\mathbf{x}$}; 
    \draw[very thick, ->, >=latex] (coordinate_origin) -- ++(45:0.707)node[above right] {$\mathbf{y}$}; 

    \def\ang1{-30}
    \path (\ang1:0.6cm) coordinate (Link1Head);
    \path (Link1Head) ++(\ang1:3cm) coordinate (Link1Tail);
    \path (Link1Head) ++(\ang1-90:0.15cm) ++(0, 1cm) coordinate (up1);
    \path (Link1Tail) ++(\ang1+90:0.15cm) ++(0, -1cm) coordinate (down1);
    
    \begin{scope}
        \clip (Link1Head) -- ++(\ang1+90:0.15cm) 
                          -- ++(\ang1:3cm) 
                          -- (Link1Tail) 
                          -- ++(\ang1-90:0.15cm)
                          -- ++(\ang1+180:3cm)
                          -- cycle;
        \fill[lightgray] (up1) rectangle (down1);
    \end{scope}
    
    % Draws base frame
    \draw[very thick, ->, >=latex] (Link1Tail) -- ++(\ang1:1cm)node[right] {$\mathbf{x_{b}}$};
    \draw[very thick, ->, >=latex] (Link1Tail) -- ++(\ang1+90:1cm)node[right] {$\mathbf{z_{b}}$};
    \draw[very thick, ->, >=latex] (Link1Tail) -- ++(\ang1+45:0.707cm)node[right] {$\mathbf{y_{b}}$};
    
    % Draws the second link
    \def\ang2{-75}
    \path (Link1Tail) ++(2, -1) coordinate (Link2Head);
    \path (Link1Tail) ++(0, -1) coordinate (Link3Head);
    \path (Link2Head) ++(\ang2:3cm) coordinate (Link2Tail);
    \path (Link2Head) ++(\ang2:1.5cm) coordinate (Link2CoM);
    \path (Link2Head) ++(\ang2+90:0.15cm) ++(-1cm, 0) coordinate (up2);
    \path (Link2Tail) ++(\ang2+90:0.15cm) ++(1, -1cm) coordinate (down2);
    \path (Link2Head) ++(\ang2+180:0.1cm) coordinate (force1head);
    \path (force1head) ++(15:2cm) coordinate (force1tail);
    \path (Link2Tail) ++(\ang2:0.1cm) coordinate (force2head);
    \path (force2head) ++(-175:2cm) coordinate (force2tail);
    
    \begin{scope}
        \clip (Link2Head) -- ++(\ang2+90:0.15cm) 
                      -- ++(\ang2:3cm) 
                      -- (Link2Tail) 
                      -- ++(\ang2-90:0.15cm)
                      -- ++(\ang2+180:3cm)
                      -- cycle;
        \fill[lightgray] (up2) rectangle (down2);
    \end{scope}
    
    % Draws base frame
    \draw[very thick, ->, >=latex] (Link2Head) -- ++(\ang2:1cm)node[right] {$\mathbf{x_{f}^{'}}$};
    \draw[very thick, ->, >=latex] (Link2Head) -- ++(\ang2+90:1cm)node[right] {$\mathbf{z_{f}^{'}}$};
    \draw[very thick, ->, >=latex] (Link2Head) -- ++(\ang2+45:0.707cm)node[right] {$\mathbf{y_{f}^{'}}$};
    
    % Draw second link dashed
    \def\ang3{-135}
    \path (Link1Tail) ++(0, -0.3) coordinate (Link3Head);
    \path (Link3Head) ++(\ang3:3cm) coordinate (Link3Tail);
    \path (Link3Head) ++(\ang3:1.5cm) coordinate (Link3CoM);

    \draw[densely dashed] (Link3Head) -- ++(\ang3+90:0.15cm)
                                      -- ++(\ang3:3cm)
                                      -- (Link3Tail) 
                                      -- ++(\ang3-90:0.15cm)
                                      -- ++(\ang3+180:3cm)
                                      -- cycle;
    
    % Draws base frame
    \draw[very thick, ->, >=latex] (Link3Head) -- ++(\ang3:1cm)node[above left] {$\mathbf{x_{f}}$};
    \draw[very thick, ->, >=latex] (Link3Head) -- ++(\ang3+90:1cm)node[below left] {$\mathbf{z_{f}}$};
    \draw[very thick, ->, >=latex] (Link3Head) -- ++(\ang3+45:0.707cm)node[below left] {$\mathbf{y_{f}}$};
    
    % Draws rotation arc
    \path (Link3CoM) ++(\ang3 + 90:0.15cm) coordinate (arc_start);
    \def\ang4{-75}
    \path (Link2CoM) ++(\ang4 + 90:-0.175cm) coordinate (arc_end);
    \draw[very thick, ->, >=latex] (arc_start) to[bend right] (arc_end);

\end{tikzpicture}
    \caption{This graph shows how the base frame and follower frame is defined, where we have the base frame defined as $\{x_b, y_b, z_b\}$ and the follower frame defined as $\{x_f, y_f, z_f\}$. Note that when calculating forces $F = k_p\Delta{x} + k_d\dot{x}$, the $\Delta{x}$ is the translation of the follower frame relative to the base frame given in world frame coordinates, the same goes to $\Delta{y}$ and $\Delta{z}$. And for $\dot{x}$ it is the velocity of the follower frame relative to the base frame given in world frame coordinates. The orientation of the links are given in quaternions, therefore the relative rotation between the two links are also given in quaternions and one can use $\omega = 2\dot{q}q^*$ to get the relative angular velocity, where $q$ is the quaterion that represents the relative rotation, and can obtain the relative angular acceleration using $\alpha = 2(\ddot{q}q^* - (\dot{q}q^*)^2)$. We can transform the quaternions to a set of Z-Y-X Euler angles and using the Euler angles we can get the corresponding torques as $\tau = k_{p\theta}\Delta{\theta} + k_{d\theta}\dot{\theta}$.}
    \label{fig:frames}
\end{figure}
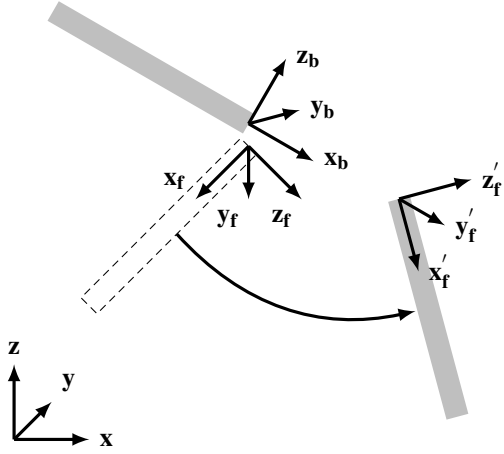
the right part of the equation is the sum of all external torques, which include pure torques and the torques generated by forces, which can also be obtained by 
\begin{equation}
	\sum{\tau} = I'\alpha + \omega\times(I'\omega)
	\label{eq:sum2}
\end{equation}
with $I' = RIR'$, where $I$ is the inertia tensor given in a frame that is rigidly connected to the link, $R$ being the rotation matrix between the body frame of the current link and the inertial frame and $\omega$ being the angular velocity of the body frame of the current link relative to the world frame given in world frame coordinates. Note that here except for $I$ everything else are given in world frame coordinates, therefore $F$ and $\tau$ are different from the forces and torques given in equation \ref{eq:f} and \ref{eq:t}. To connect the two we have to apply a rotation matrix, $F_{\rm world} = R\cdot F_{\rm body}$ and $\tau_{\rm world} = R\cdot\tau_{\rm body}$. After combining equation \ref{eq:sum1} and \ref{eq:sum2} we can predict the movement of any robotic link and see its connection with the forces and torques that are applied to it.

\subsection{Reconstruction Algorithm}

Using the equation above we can shuffle its structure so that we can perform linear regression in reconstructing the joint parameters. We can transform equation \ref{eq:sum1} and \ref{eq:sum2} into 
\begin{equation}
	\sum{\tau_F} + \sum{\tau_\tau} = I'\alpha_s + \omega_s\times(I'\omega_s)
	\label{eq:pre}
\end{equation}
with 
\begin{align*}
	\sum{\tau_F} &= (R_1\cdot r_{b1})\times(R_1\cdot F_{b1}) - (R_2\cdot r_{b2})\times(R_2\cdot F_{b2})\\
	\sum{\tau_\tau} &= R_1\cdot\tau_{b1} - R_2\cdot\tau_{b2}
\end{align*}
we also can get $F_{bi}$, $\tau_{bi}$ and $r_{bi}$ as
\begin{align*}
	F_{bi} &= \begin{bmatrix}
			\Delta{x_i} & \dot{x_i} & 0 & 0 & 0 & 0\\
			0 & 0 & \Delta{y_i} & \dot{y_i} & 0 & 0\\
			0 & 0 & 0 & 0 & \Delta{z_i} & \dot{z_i}
		\end{bmatrix}\begin{bmatrix}
			k_{pxi}\\
			k_{dxi}\\
			k_{pyi}\\
			k_{dyi}\\
			k_{pzi}\\
			k_{dzi}
		\end{bmatrix}\\
	\tau_{bi} &= \begin{bmatrix}
			\Delta\theta_{xi} & \dot{\theta_{xi}} & 0 & 0 & 0 & 0\\
			0 & 0 & \Delta\theta_{yi} & \dot{\theta_{yi}} & 0 & 0\\
			0 & 0 & 0 & 0 & \Delta\theta_{zi} & \dot{\theta_{zi}}
		\end{bmatrix}\begin{bmatrix}
			k_{p\theta_{xi}}\\
			k_{d\theta_{xi}}\\
			k_{p\theta_{yi}}\\
			k_{d\theta_{yi}}\\
			k_{p\theta_{zi}}\\
			k_{d\theta_{zi}}
		\end{bmatrix}\\
	r_{bi} &= \begin{bmatrix}
			r_{ix}\\
			r_{iy}\\
			r_{iz}
		\end{bmatrix}
\end{align*}
with $i = 1, 2$, where $1$ denotes the head joint and $2$ denotes the tail joint. Also we lets have $R_1$ and $R_2$ as
\begin{align*}
	R_1 &= \begin{bmatrix}
			m_{11} & m_{12} & m_{13}\\
			m_{21} & m_{32} & m_{33}\\
			m_{31} & m_{32} & m_{33}\\
		\end{bmatrix}\\
	R_2 &= \begin{bmatrix}
			n_{11} & n_{12} & n_{13}\\
			n_{21} & n_{32} & n_{33}\\
			n_{31} & n_{32} & n_{33}\\
		\end{bmatrix}
\end{align*}
where we have $R_1$ denoting the rotation matrix of the link that is rigidly connected to the base frame of the head joint, and $R_2$ being the rotation matrix of the link that is rigidly connected to the base frame of the tail joint, which is typically the current link. The linear regression is in the form of
\begin{center}
	$A\cdot c = b \rightarrow c = (A^T\cdot A)^{-1}\cdot A^T\cdot b$
\end{center}
therefore, we need to restructure equation \ref{eq:pre} into a linear-regression-able form. We can get the $A$ matrix as,
\begin{center}
	$A = \begin{bmatrix}
		A_1 & A_2 & A_3
	\end{bmatrix}$
\end{center}
where we have
\begin{align*}
	A_1 &= A_{11} .* A_{12}\\
	A_2 &= A_{21} .* A_{22}\\
	A_3 &= \begin{bmatrix}
			\theta_1 .* \begin{bmatrix}
			R_1 & R_1
		\end{bmatrix} & \theta_2 .* \begin{bmatrix}
			R_2 & R_2
		\end{bmatrix}
		\end{bmatrix}
\end{align*}
with
\begin{align*}
	A_{11} =& \begin{bmatrix}
			m_{31} & m_{32} & m_{33}\\
			m_{11} & m_{12} & m_{13}\\
			m_{21} & m_{32} & m_{33}\\
		\end{bmatrix}.*\begin{bmatrix}
			r_{1y}\\
			r_{1z}\\
			r_{1x}
		\end{bmatrix}\\
		-& \begin{bmatrix}
			m_{21} & m_{22} & m_{23}\\
			m_{31} & m_{32} & m_{33}\\
			m_{11} & m_{12} & m_{13}\\
		\end{bmatrix}.*\begin{bmatrix}
			r_{1z}\\
			r_{1x}\\
			r_{1y}
		\end{bmatrix}\\
	A_{12} =& \begin{bmatrix}
			\Delta_{x1} & \Delta_{y1} & \Delta_{z1} & v_{x1} & v_{y1} & v_{z1} 
		\end{bmatrix}\\
	A_{21} =& \begin{bmatrix}
			n_{31} & n_{32} & n_{33}\\
			n_{11} & n_{12} & n_{13}\\
			n_{21} & n_{32} & n_{33}\\
		\end{bmatrix}.*\begin{bmatrix}
			r_{2y}\\
			r_{2z}\\
			r_{2x}
		\end{bmatrix}\\ 
		-& \begin{bmatrix}
			n_{21} & n_{22} & n_{23}\\
			n_{31} & n_{32} & n_{33}\\
			n_{11} & n_{12} & n_{13}\\
		\end{bmatrix}.*\begin{bmatrix}
			r_{2z}\\
			r_{2x}\\
			r_{2y}
		\end{bmatrix}\\
	A_{22} =& \begin{bmatrix}
			\Delta_{x2} & \Delta_{y2} & \Delta_{z2} & v_{x2} & v_{y2} & v_{z2} 
		\end{bmatrix}\\
	\theta_1 =& \begin{bmatrix}
			\theta_{x1} &
			\theta_{y1} &
			\theta_{z1} &
			\dot\theta_{x1} &
			\dot\theta_{y1} &
			\dot\theta_{z1}
		\end{bmatrix}\\
	\theta_2 =& \begin{bmatrix}
			\theta_{x2} &
			\theta_{y2} &
			\theta_{z2} &
			\dot\theta_{x2} &
			\dot\theta_{y2} &
			\dot\theta_{z2}
		\end{bmatrix}
\end{align*}
where we have $.*$ as element wise multiplication. Also we have
\begin{center}
	$b = I'\alpha_s + \omega_s\times(I'\omega_s)$
\end{center}
and
\begin{center}
	$c = \begin{bmatrix}
		K_1\\
		K_2\\
		K_{\theta_1}\\
		K_{\theta_2}
	\end{bmatrix}$
\end{center}
with
\begin{center}
	$K_i = \begin{bmatrix}
			k_{pxi}\\
			k_{dxi}\\
			k_{pyi}\\
			k_{dyi}\\
			k_{pzi}\\
			k_{dzi}
		\end{bmatrix}$\ \ \ $K_{\theta_i} = \begin{bmatrix}
			k_{p\theta_{xi}}\\
			k_{d\theta_{xi}}\\
			k_{p\theta_{yi}}\\
			k_{d\theta_{yi}}\\
			k_{p\theta_{zi}}\\
			k_{d\theta_{zi}}
		\end{bmatrix}$
\end{center}
where $i = 1, 2$, with denotes the head and tail joint respectively.
\begin{figure}
    \centering
    \input{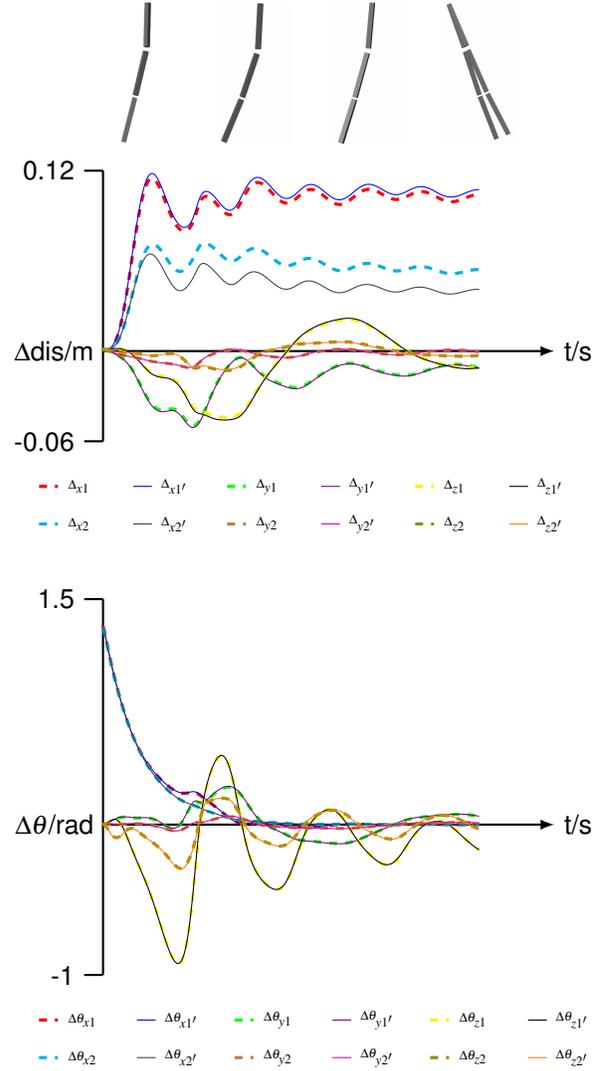}
    \caption{
    This graph shows the performance of the proposed reconstruction algorithm using simulated data. Note that we are in interest of the joints that connects the middle link to the upper and lower link. The upper figure shows the evolution of relative translation inside the joints over time, where the subscript $x1$ denotes the relative translation along the $x$-axis in the body frame for upper joint, and $x2$ denotes the same for the lower joint (note that both joints are on the second link when counting from up to down), and $x1'$ denotes the reconstructed version. The same goes for the lower graph where the evolution of relative rotation is shown. The comparison in the top shows the different poses of the three link system at time 1s, 4s, 7s and 10s.
    }
    \label{fig:sim_result}
\end{figure}
\subsection{Simulation Validation}
\begin{figure*}[t]
    \centering
    \includegraphics[width=\textwidth]{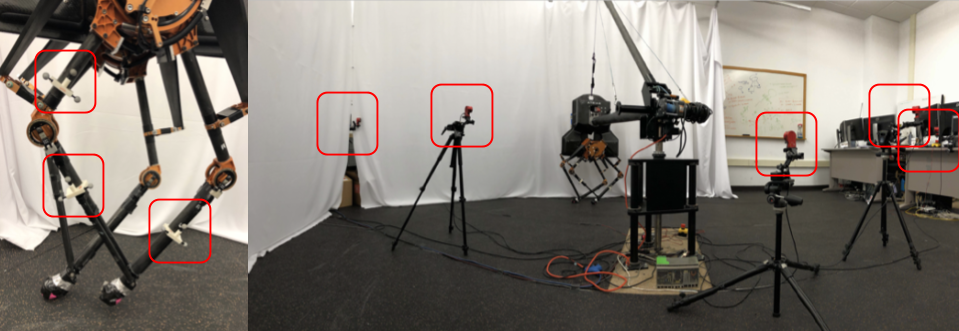}
    \caption{On the left we can see the markers attached to the leg links, we currently have 3 marker mounts that each provides a coordinate frame that is rigidly attached to the robot. On the right the camera positions are shown, the robot will hop at place which gives the best capture view for the motion capture system.}
    \label{fig:setup}
\end{figure*}
We built a serial-link-open-chain model in simulation to test the validity of the aforementioned reconstruction algorithm. Simulink is a graphical programming environment for modeling, simulating and analyzing multidomain dynamic systems, which is the ideal candidate for our task. Using the Simscape Multibody environment we built a three-link pendulum model where each link is connected through a six-DOF joint, for simulation tolerance we have a relative tolerance of $1e^{-6}$ and an absolute tolerance of $1e^{-8}$, ode15 is the integration solver in use and the sample time is $1/120s$. We obtained all of the measurements required to perform the reconstruction from the simulation sensory data, using such data we got the reconstructed joint parameters and compared them with the pre-defined joint parameters that are used to construct the model. We noticed that they are approximately the same. To make the simulation setup more alike to the hardware experiment, we use only the position and orientation data (given in quaternions) of the links in interest as the input of the algorithm. The velocity and acceleration data both linear and rotational is obtained from the position data and orientation data. After exploring with multiple algorithms we used what is proposed in \cite{Sittel2013ComputingVA}. The simulation result is shown in Fig \ref{fig:sim_result}. We can see that in the top we have poses of the three link system, the first three shows little difference between the poses from the simulation and the reconstructed version, however for the fourth one the difference is more significant. The reason for this is the first three are closer to the neutral position, and the difference grows as the three link system deviates from the neutral position.
\section{EXPERIMENTAL EVALUATION}

\subsection{Hardware Setup}
Here we introduce the OptiTrack motion capture system and the ATRIAS bipedal robot as the experiment platform. The OptiTrack system is developed by Natural Point, we use an array of five Flex 13 cameras which are mounted to approximately 1.5 meters high and captures frames at 120Hz, which provides full coverage for the leg links. The ATRIAS bipedal robot is a human-scale bipedal robot designed and built by the Oregon State University Dynamic Robotics Lab, which has been demonstrated to have agile walking, running and hopping gaits. To obtain the position and orientation of the leg links we added 3D printed marker mounts, which gives a coordinate system that is rigidly attached to the link, we use this coordinate system to estimate the orientation of the leg links. The whole experiment setup is shown in figure \ref{fig:setup}. The inertia matrix and the center-of-mass location of the leg links are determined from the original design drawing.

\subsection{Method}
\begin{algorithm}[t]
    \caption{Get Relative Translation Between Two Links}\label{alg:static}
    \begin{algorithmic}[1]
    \State Obtain linkage rotation data $\mathbf{R_1}$ and $\mathbf{R_2}$
    \State Find a tracking point on each link $x_1$ and $x_2$ in world frame coordinates
    \State Get the distance of the tracking point and the link-joint connection point in body frame coordinates $r_1$ and $r_2$
    \State Get the distance between the tracking points in world frame coordinates $\mathbf{X} = x_1 - x_2$
    \State Assume no relative translation the distance between tracking points is $\mathbf{X}_{\rm img} = -R_1r_1 - R_2r_2$ in world frame coordinates
    \State The joint relative translation in world frame coordinates can be obtained from $\mathbf{X_{\rm rel}} = \mathbf{X}_{\rm img} - \mathbf{X}$
    \State The joint relative translation in base frame coordinates can be obtained from $\mathbf{X}_{\rm relbody} = R_1^T\mathbf{X}_{\rm rel}$
    \end{algorithmic}
\end{algorithm}

\begin{figure}[t]
    \centering
    \input{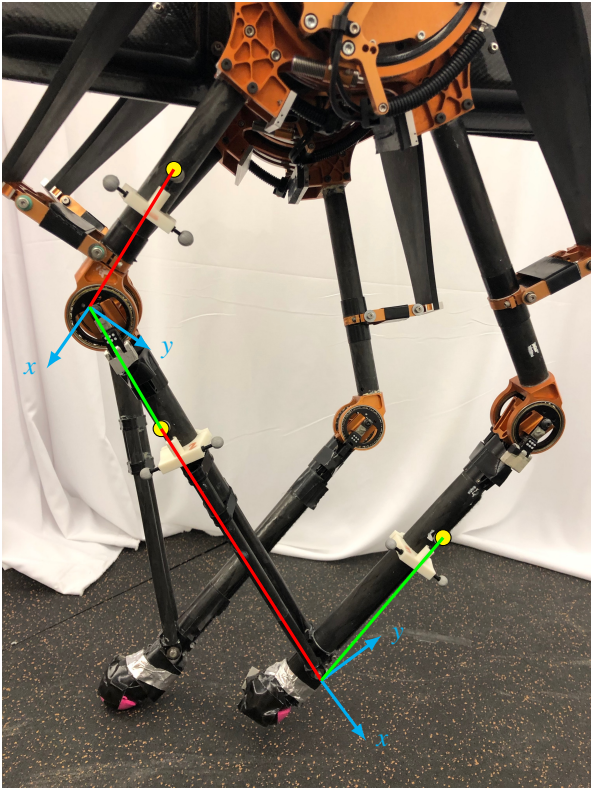}
    \caption{This figure shows the requiring measurement for estimating the relative translation and relative rotation among the joints. The cyan frame on each joint represents the body frame where the $x$-axis points along the red line, and the $y$-axis is perpendicular to the red line which is the same frame that is define by the three markers on each link. Using a pair of red and green measurements along with algorithm 1 we can get the relative translation.}
    \label{fig:markers}
\end{figure}
\begin{figure}[t]
    \centering
    \begin{subfigure}[b]{0.4\textwidth}
        \includegraphics[width=\textwidth]{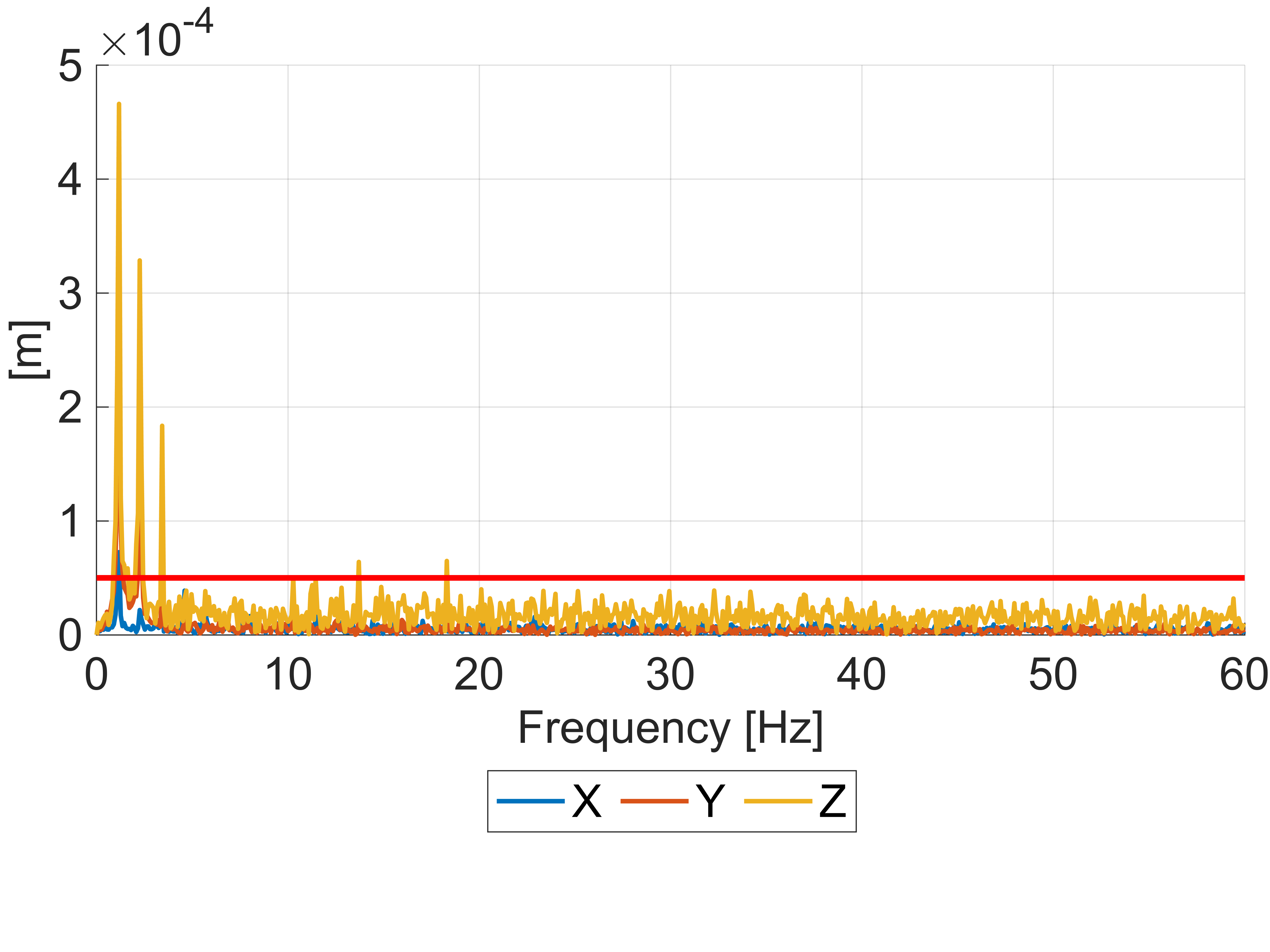}
        \caption{}
        \label{fig:cutoff}
    \end{subfigure}
    \begin{subfigure}[b]{0.49\textwidth}
        \includegraphics[width=\textwidth]{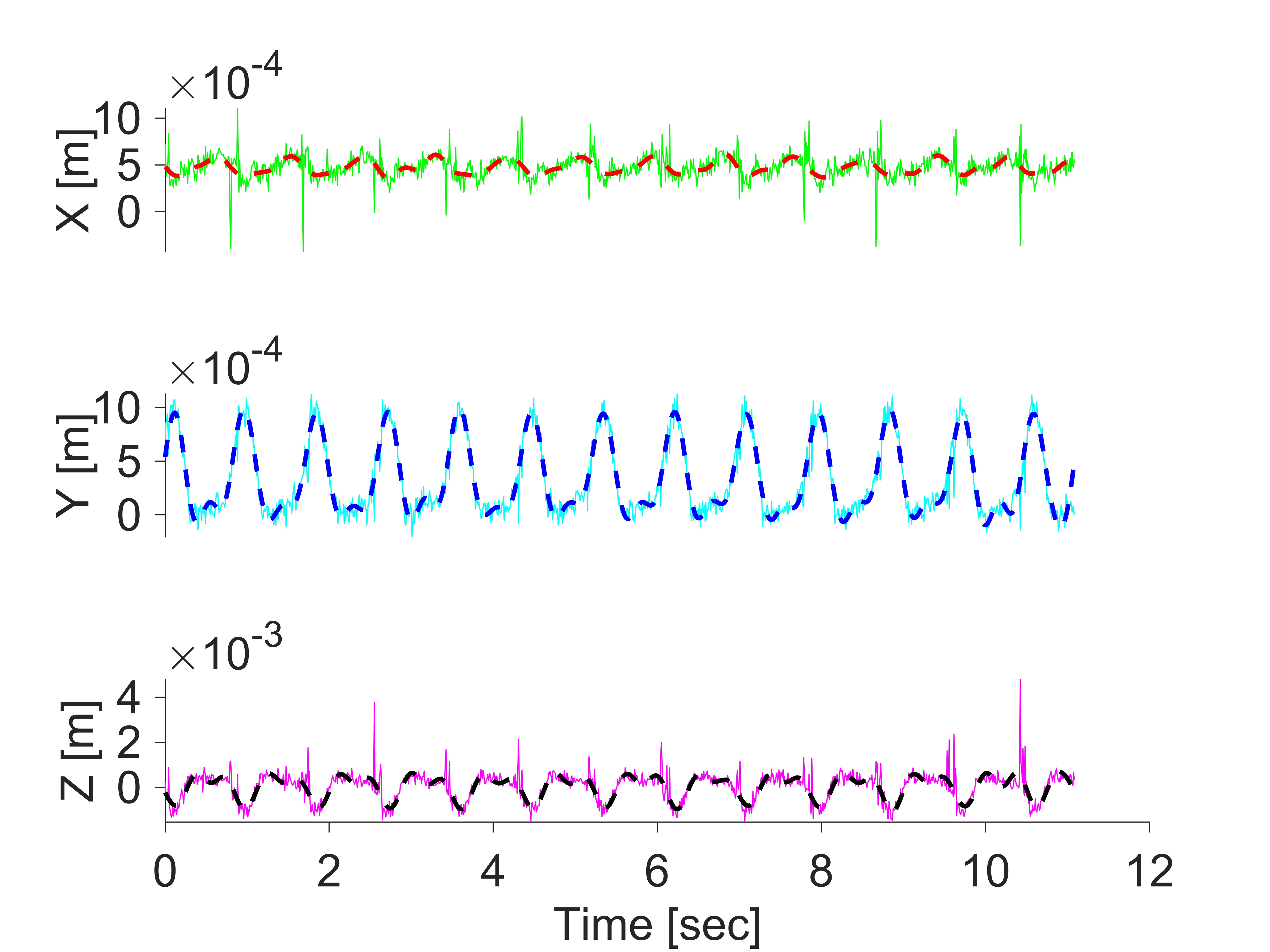}
        \caption{}
        \label{fig:filtResult}
    \end{subfigure}
    \caption{This figure shows the filtering process, figure \ref{fig:cutoff} shows the frequency information of the relative translation for the upper joint, which is shown in figure \ref{fig:markers}. From obtained white noise frequency information we can draw a line that indicates whether the corresponding frequency component can be considered as white noise. Using this information the cutoff frequency is determined to be 5Hz, and the post-filtering result is shown in figure \ref{fig:filtResult}, where the dashed lines are the post-filtered data and the solid lines are the pre-filtered data. We can see that after filtering the noise has been greatly reduced while the loss of information is minimal.}
\end{figure}
During the experiment ATRIAS performs a low speed running gait \cite{DBLP:journals/ral/MartinWG17} while an external force limits the forward motion, the result is a gait that is similar to hopping at place. After obtaining the motion capture data we extract the relative translation and relative rotation of each joint. After obtaining the relative translation and rotation for both the knee and ankle joints we can put all the measurements into the reconstruction algorithm mentioned in the previous section and obtain the corresponding joint coefficients. Note that in the current setting the reconstruction is only for the middle link shown in figure \ref{fig:markers}, this is because the forced acting on the middle link is only produced by the knee and ankle joints. Whereas the other links may have forces and torques produced by actuation or interaction with the external environment. The hip joint is relatively rigid thus the displacement inside the joint is much more difficult to measure using a motion capture system. Also we lack a method for measuring the ground reaction force for the link that touched the ground.

From the markers we can obtain the orientation of the link using the connecting lines of each marker as the axis of the body frame. However using this method we usually cannot obtain an orthonormal matrix, we then find the nearest orthonormal matrix in terms of Frobenius norm, and replace that as the rotation matrix. Then it is trivial to obtain the relative rotation: after obtaining the orientation of each leg link that are connected to a joint $R_1$ and $R_2$, the relative rotation is obtained by $R_{\rm rel} = R_1^TR_2$. To obtain the relative translation we use algorithm \ref{alg:static}. The idea for getting the relative translation is to compare the difference between two points when there is relative translation in the joints and when there is not. The difference between the two will be the relative translation, which can be shown in figure \ref{fig:markers}.

\begin{figure*}[t]
    \centering
    \includegraphics[width=\textwidth]{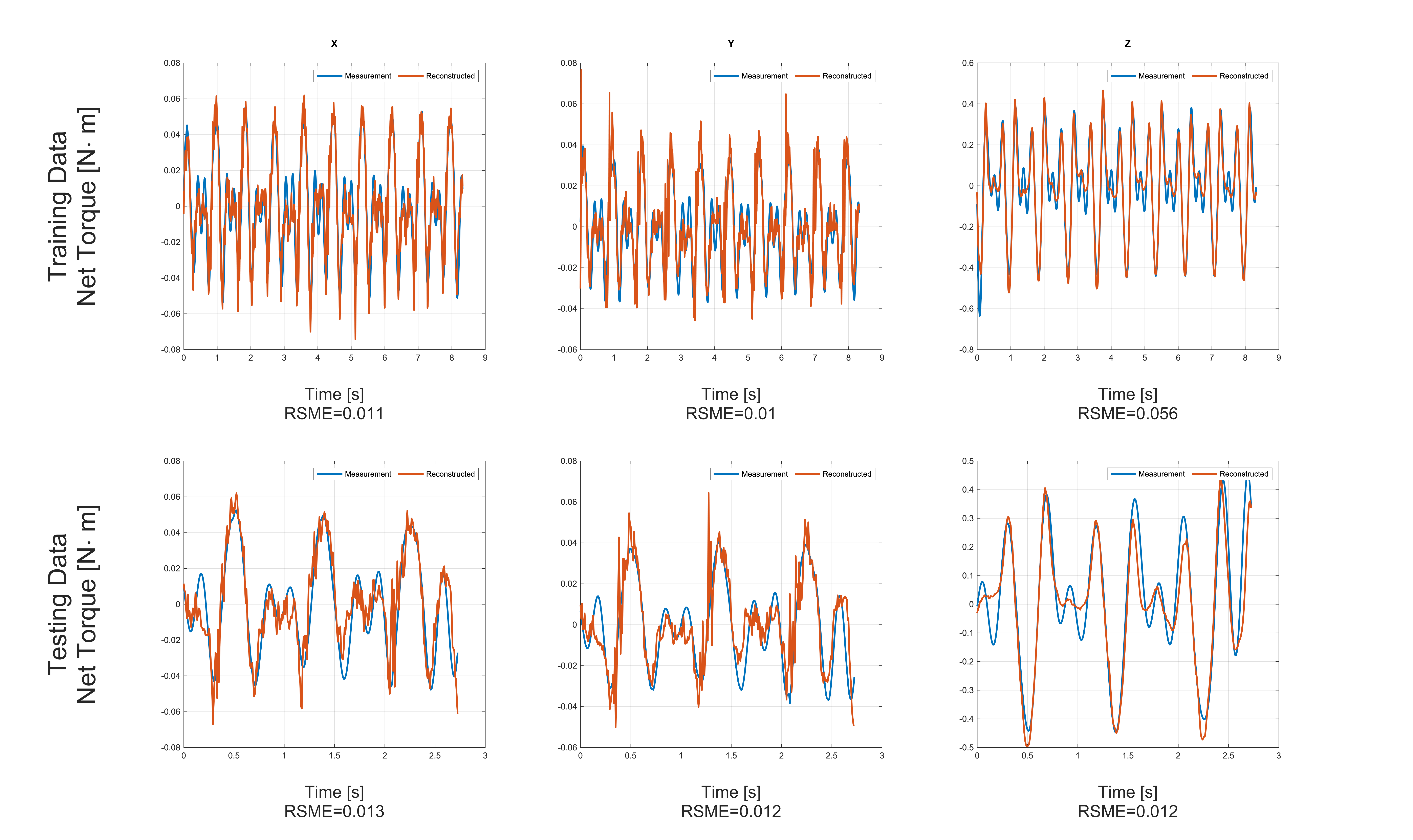}
    \caption{This figure shows the reconstruction result using data obtained from the ATRIAS biped. The data can be separated into a training set and a testing set. We obtain the joint coefficients from the training set and use them with the testing set measurements to get the corresponding net torque estimations. We see that both the training set and the testing set fits the data well which can also be indicated by the RMSE error.}
    \label{fig:result}
\end{figure*}

\begin{table*}[t]

\centering
\begin{tabular}{@{}cccccccccccc@{}}
\toprule
$k_{px}$ & $k_{py}$ & $k_{pz}$ & $k_{dx}$ & $k_{dy}$ & $k_{dz}$ &
 $k_{p\theta x}$ & $k_{p\theta y}$ & $k_{p\theta z}$ & $k_{d\theta x}$ & 
 $k_{d\theta y}$ & $k_{d\theta z}$ \\ 
\midrule
 1.76e3 & 2.65e4 & 2.15e2 & 5.53e2 & -18.26 & 7.78 & 1.58 & -2.15 & -1.06 & -0.06 & -0.01 & 0.02 \\
 6.60e-7 & -65.12 & -9.05 & -3.54e-9 & 22.72 & 0.25 & 0.52 & 0.80 & 2.94 & 0.01 & 0.04 & -0.02 \\
\bottomrule
\end{tabular}
\caption{This table gives the reconstructed joint coefficients, the first row corresponds to the upper joint shown in figure \ref{fig:markers} and the second row corresponds to the lower joint.}
\label{tab:joint}
\end{table*}

Before inputting the measurements into the reconstruction algorithm filtering is required. High frequency noise in the data can be magnified after differentiating the position and orientation data into both linear and angular velocity and acceleration. Filtering requires a cutoff frequency given that we want the data to go through a lowpass filter. This cutoff frequency is determined by comparing the power of each frequency of the data where the robot is moving and data where the robot is standing still, which is shown in figure \ref{fig:cutoff}. When ATRIAS is standing still the relative translation should be a fix value, and the Fourier transform would show the power of the white noise. Assuming that the non-white-noise information should have a much higher power than white noise we can then determine the cutoff frequency by a simple criteria: the power of a frequency that is in the region of the white noise should be filtered out. 

After get the getting the filtered position and orientation data we need to get the corresponding velocity and acceleration for both the translational and rotational components. We do so by using the sliding-window regression algorithm proposed in \cite{Sittel2013ComputingVA}. Note that the orientation is first represented using quaternions $\mathbf{q}$ and after passing them into the sliding-window regression algorithm what we obtain is the first and second derivative of the quaternions $\dot{q}$ and $\ddot{q}$. The corresponding angular velocity and acceleration can be obtained as
\begin{align*}
    \omega &= 2\dot{q}q^*;\\
    \alpha &= 2(\ddot{q}q^* - (\dot{q}q^*)^2)
\end{align*}
where $q^*$ is the conjugate quaternion of $q$, $\omega$ is the angular velocity and $\alpha$ is the angular acceleration.

The moment of inertia, the link mass, and the distances between the markers and the joints are obtained using a combination of measurements from the SolidWorks model and on ATRIAS. Which can be inaccurate due to previous repairs and measurement error. To get a more accurate measurement we added a bias term for each measurement in the regression to compensate. However, for the link mass and moment of inertia we would need to disassemble ATRIAS to perform system identification, therefore we rely on the SolidWorks model for these measurements.

\subsection{Results}
Using the measurements obtained from the motion capture system and other sources as mentioned in the previous sections we can obtain a set of joint coefficients. To validate the accuracy of the reconstructed joint coefficients we compare the measured net torque and the net torque reconstructed using the obtained joint coefficients the result is shown in figure \ref{fig:result}. As we can see the reconstruction has a relatively low root mean squared error (RSME), which indicates that the reconstructed net torque can capture the real motion. This result also indicates the joint model of the robot and the joint coefficient changed little over time using previously obtained joint model and joint coefficients we can online monitor the torques and forces in the unactuated joints. The reconstructed joint coefficients are shown in table \ref{tab:joint}. As we can see the joint coefficients have negative terms which does not align with our joint model, for a more detailed discussion regarding this please see the discussion section. 
\section{DISCUSSIONS}
\begin{figure}[t]
    \centering
    \includegraphics[width=0.49\textwidth]{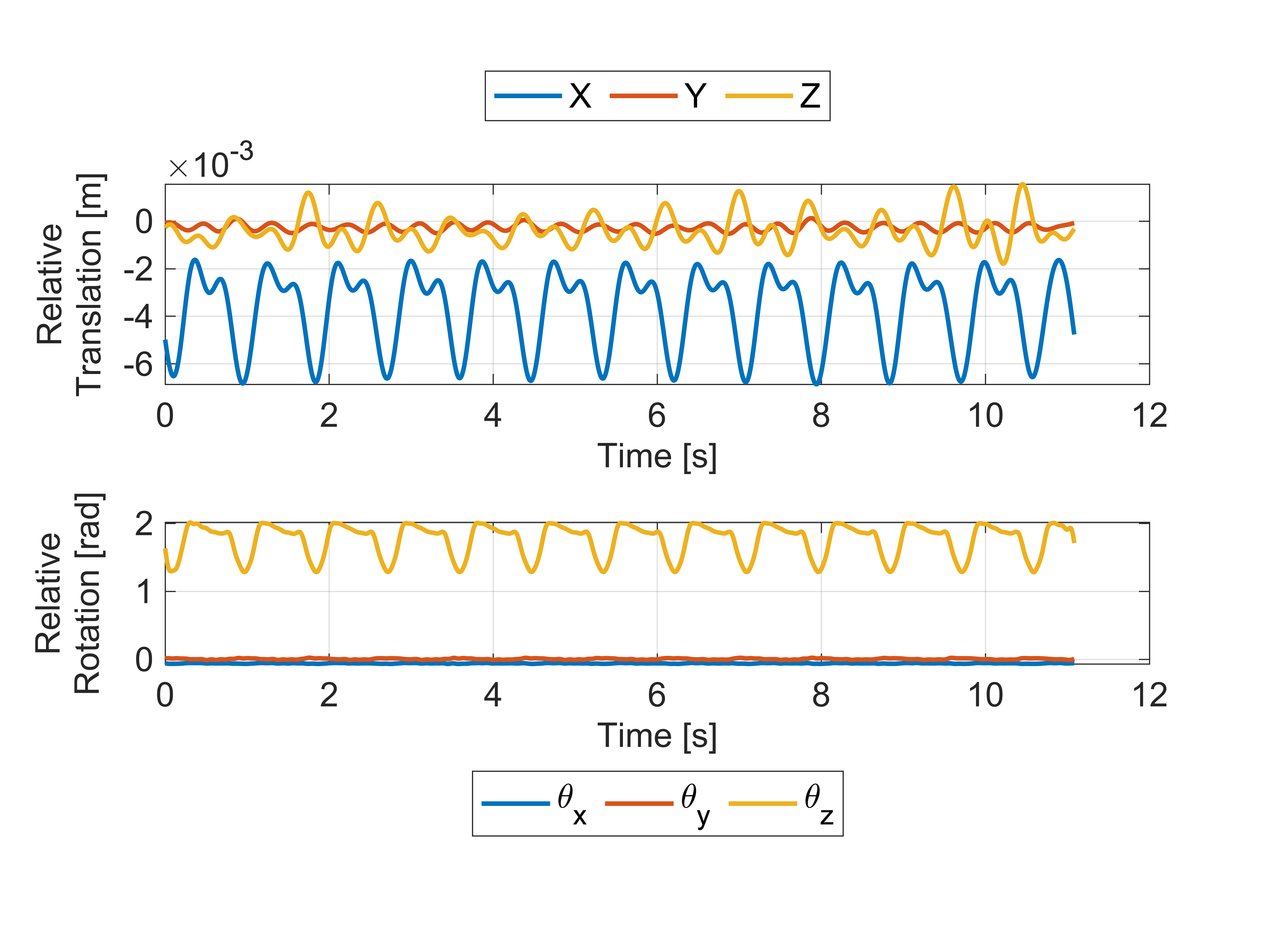}
    \caption{This figure shown the relative translation and relative rotation of the lower joint shown in figure \ref{fig:markers}. We can observe that the displacement on the $x$-axis deviates from 0 a lot. Also the magnitude of the relative rotation are about $10^{-1}$, while the magnitude of the relative translation are amongst $10^{-3}$.}
    \label{fig:measure}
\end{figure}

This work focuses on identifying the kinematic structure of legged robots adaptively. We model each joint as a 6 DOF first-order spring-damper system, where the joint coefficients can be reconstructed using the method described in section II. From the results we can see that the reconstruction algorithm can build a meaningful model of the system which is able to predict the net torque using only partial measurements from the motion capture system. Knowledge from previous work on ATRIAS shows that during high impact motions such as running and hopping the knee joint experiences out-of-plane bending which leads to fatal failures for the system. This observation is confirmed in this work from the obtained motion capture data. What can also be observed is that this deformation can be found in all DOFs of the joints. Which consolidates our belief that the knee and ankle joints of ATRIAS can be better modeled using a 6 DOF joint model. This remainder of this section is structured as follows, first we will elaborate on the motion observations, then we will discuss the effectiveness of our current model, finally we will talk about the limits of the current model and possible future directions. 

\subsection{Observed Motion}
From figure \ref{fig:result} and the motion capture data we can see both the knee and ankle joints have movements in previously considered rigid DOFs. Previous experiments have observed out-of-plane movement at the knee joint, which is magnified by the non-perpendicular placement of the leg links with respect to the ground during walking and running. In this work this observation is confirmed by observing nonzero relative rotation on all three DOFs of the knee joint, though the motion is small in scale. The movement is smaller compared to what is observed in previous works which is mainly due to gait selection. What is worth pointing out is the deformation in the lower joint, which is shown in figure \ref{fig:measure}. The relative translation among the $x$-axis is larger in range compared to the other axes, this phenomenon is due to the mechanical design of ATRIAS. Applying force by hand we can move slightly along the $x$-axis for the ankle joint, while the same amount of movement cannot be achieved on any other degrees of freedom. 

\subsection{Joint Model}
\begin{figure}
    \centering
    \includegraphics[width=0.49\textwidth]{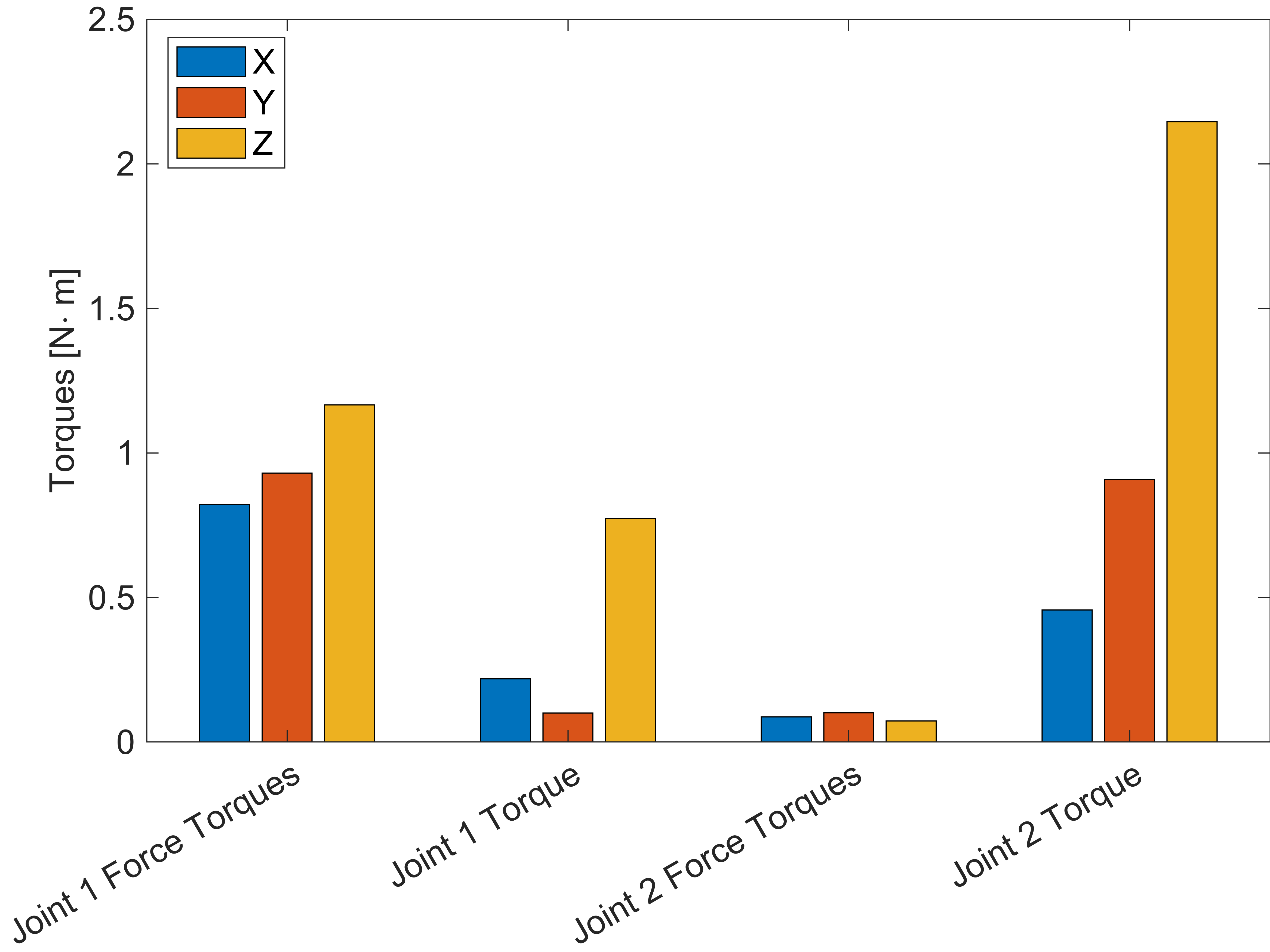}
    \caption{This figure shows the range of torques generated by both the translational and rotational spring dampers of each joint. Here the force torques represent the torques generated by force which can be calculated as $r\times F$.}
    \label{fig:forces}
\end{figure}
The current joint model is a 6 DOF joint model, which has a spring damper system on each DOF to govern its movement. Using the reconstruction algorithm in section II we can obtain the corresponding joint coefficients using motion capture data which are shown in table \ref{tab:joint}. And then we reconstructed the torques produced by joint deformation, which is shown in figure \ref{fig:forces}. We can see that the $k_p$ and $k_d$ terms for the upper joint are relatively large and for the lower joint they are much smaller. This indicates the upper joint is more rigid in the translational DOF. This can also be verified by inspecting the joints themselves. The $k_{p\theta}$ and $k_{d\theta}$ terms are small due to the inaccuracies in relative rotation measurements, which has a much larger magnitude when compared with translation measurements. Another anomaly is that some of the $k_p$ and $k_d$ terms are negative, where the current model assumes positivity for all of the joint coefficients. A possible explanation for this would be due to inaccurate estimations for the neutral points of the joints along with using a non-representative joint model. From Hooke's law we know the force produced by a spring is proportional with the displacement relative to the neutral position. This explains why the $k_{px}$ and $k_dx$ term are significantly smaller for the lower joint in figure \ref{fig:markers} compared to other force related joint coefficients. The measured displacements on the $x$-axis for the second joint ranges between -2 and -8 millimeters, while all the other force related displacements are in the sub-millimeter regime, which is shown in figure \ref{fig:filtResult}. Additionally, translational displacements are of the magnitude of $10^{-4}$ while the rotational displacements given in radians are among $10^{-1}$. This also may be the reason which leads to disproportion in magnitude between force-related and torque-related joint coefficients.

\subsection{Limitations \& Future Directions}
There are a few limitations to the current approach. This work models the joints as a connection between links which uses a spring-damper system to govern the movement on all 6 DOF. In reality, most springs acts linearly only in a certain region, the overall model is nonlinear. We tried fitting a third order and fifth order model to the system, but the results did not show a significant improvement. Though by looking into the reconstruction results we observed higher order models has the ability to fit high frequency components in the data better, thus slightly reducing the RMSE. However, the use of these models lacks physical intuition for its mechanism, further research is required before commiting to these models. Previous work on deformation in elastic materials uses a nonlinear elastic model to describe the force-deformation relationship. Work in neuromuscular control \cite{6611284}, \cite{HAEUFLE20141531} have been using many complex versions of spring-damper systems to model forces in muscles. Since the springs are inherently nonlinear experimenting with nonlinear models might produce promising results.

The current validation method is to separate the data into two parts: one used for training and the other used for testing (validation). The RSME of the training set shows how well the reconstruction algorithm can fit the data, while the testing set shows whether the reconstructed joint coefficients overfits to the training set. This provides a metric to evaluate the performance of the reconstruction algorithm. However, this only validates the ability for finding a model which explains the net torque of the link. A more desirable validation method would also give you the accuracy for the reconstruction of the torques and forces for a specific joint (the knee and ankle joint). To achieve this we need to measure the ground truth of the forces and torques for each separate joint. We can achieve this by measuring the ground reaction force (GRF) using a force plate. Once the GRF is known we can perform inverse kinematics to get the corresponding torques and forces on the knee and ankle joint. Then we can compare it with the reconstructed force and torques to get a better understanding for the accuracy of the reconstruction algorithm. Which also will give us a method for validating the accuracy of the joint model.

% where the force and torque on each DOF can be described as,
% \begin{align*}
%     F_{3i} &= k_{p1}\Delta_i + k_{p2}\Delta_i^3 + k_{d1}\dot{v}_i + k_{d2}\dot{v}_i^3\\
%     \tau_{3i} &= k_{pt1}\Delta\theta_i + k_{p2}\Delta\theta_i^3 + k_{d1}\dot{\theta_i} + k_{d2}\dot{\theta_i}^3\\
%     F_{5i} &= k_{p1}\Delta_i + k_{p2}\Delta_i^3 + k_{p3}\Delta_i^5 + k_{d1}\dot{v}_i + k_{d2}\dot{v}_i^3 + k_{d3}\dot{v}_i^5\\
%     \tau_{5i} &= k_{pt1}\Delta\theta_i + k_{p2}\Delta\theta_i^3 + k_{p3}\Delta\theta_i^5 + k_{d1}\dot{\theta_i} + k_{d2}\dot{\theta_i}^3 + k_{d3}\dot{\theta_i}^5\\
% \end{align*}
% where $i = \{x, y, z\}$. 

% \begin{table}
% \centering
% \caption{RMSE for different joint models}
% \begin{tabular}{@{}cccc@{}}
% \toprule
%  Data                                & First Order & Third Order & Fifth Order \\ \midrule
%  Train $\displaystyle\Sigma{\tau_x}$ & 0.0107      & 0.0095      & 0.0093 \\
%  Test $\displaystyle\Sigma{\tau_x}$  & 0.0113      & 0.0119      & 0.0108 \\
%  Train $\displaystyle\Sigma{\tau_y}$ & 0.0087      & 0.0071      & 0.0075 \\
%  Test $\displaystyle\Sigma{\tau_y}$  & 0.0099      & 0.0088      & 0.0092 \\
%  Train $\displaystyle\Sigma{\tau_z}$ & 0.0697      & 0.0572      & 0.0574 \\ 
%  Test $\displaystyle\Sigma{\tau_z}$  & 0.0907      & 0.0750      & 0.0640 \\ \bottomrule
% \end{tabular}
% \label{tab:models}
% \end{table}
\section{CONCLUSIONS}
We present work to automate identification for humanoid kinematic structure which takes the possible deformation into consideration, and constructed an algorithm to reconstruct the kinematic structure of the robotic system from obtained joint parameters. Initially in simulation the reconstruction algorithm is validated. Then we experimented on the ATRIAS biped and collected the data using an array of OptiTrack motion capture cameras. We showed that our method can fit a model to the data which is able to reconstruct the net torque accurately. However, the current method presents some limitations. Firstly, the calibration depends largely on measurements which can be improved by using the motion capture system to assist the calibration, also adding a bias term into the regression would also aid to this. Secondly, the current method uses a first order linear joint model which does not capture the nonlinearity in the system well, this can be improved by using a higher order linear model or even nonlinear joint models. Additionally, the current validation method could only only evaluate the accuracy for reconstructing the net torque generated by both the knee and ankle joint, while knowing the accuracy for each separate joint would be more ideal. This can be improved by measuring the GRF.
\section{ACKNOWLEDGMENT}
I want to thank Ashwin Khadke, Timothy Kyung, Tianyu Li, Justin Macey, William Martin, Akshara Rai, Avinash Siravuru and Nitish Thatte for their support on operating ATRIAS, tuning the motion capture system and many inspiring discussions. I also want to thank Taiyan Liu for her tremendous support during my entire academic career at Carnegie Mellon University, which I am forever grateful.
\bibliographystyle{IEEEtran}
\bibliography{asme2e}

\end{document}